\documentclass[letterpaper,journal]{IEEEtran}
\usepackage{amsmath,amsfonts,amssymb}
\usepackage{algorithmic}
\usepackage{algorithm}
\usepackage{array}
\usepackage[caption=false,font=normalsize,labelfont=sf,textfont=sf]{subfig}
\usepackage{textcomp}
\usepackage{stfloats}
\usepackage{url}
\usepackage{verbatim}
\usepackage{graphicx}
\usepackage{cite}
\usepackage[table]{xcolor}
\usepackage{booktabs}
\usepackage{multirow}
\begin{document}

\title{PathAR: Structure-First Autoregressive Synthesis \\ of Multimodal Pathology Images}

\author{Yuan Zhang,
        Jiahao Xia,
        Junzhang Huang,
        Meng Wang,
        Feng Chen,
        Guanyu Yang,~\IEEEmembership{Senior Member,~IEEE},
        and Huazhu Fu,~\IEEEmembership{Senior Member,~IEEE}%
\thanks{Yuan Zhang, Jiahao Xia, Junzhang Huang, and Guanyu Yang are with the Key Laboratory of New Generation Artificial Intelligence Technology and Its Interdisciplinary Applications (Southeast University), Ministry of Education, Nanjing 210096, China (e-mail: yuanzhang\_@seu.edu.cn; xiajiahao46@gmail.com; huangjunzhang2022@163.com, yang.list@seu.edu.cn).}%
\thanks{Meng Wang is with the Centre for Innovation and Precision Eye Health, Yong Loo Lin School of Medicine, National University of Singapore, Singapore 119228, Singapore, and the Department of Ophthalmology, Yong Loo Lin School of Medicine, National University of Singapore, Singapore 119228, Singapore (e-mail: wang.m@nus.edu.sg).}%
\thanks{Feng Chen is with the Department of Biostatistics, Center for Global Health, School of Public Health, Nanjing Medical University, Nanjing 211166, China (e-mail: fengchen@njmu.edu.cn).}%
\thanks{Huazhu Fu is with the Institute of High-Performance Computing, Agency for Science, Technology and Research, Singapore 138632 (e-mail: hzfu@ieee.org).}%
\thanks{Corresponding author: Guanyu Yang and Huazhu Fu.}}



\maketitle

\begin{abstract}
Data scarcity in multimodal pathology motivates unified generative models that synthesize modality-specific appearance while preserving anatomically coherent structure. 
Although modalities differ in appearance statistics, morphological structures such as cellular topology and tissue boundaries are largely preserved across acquisition protocols. 
However, existing methods often model these factors within a homogeneous token stream, implicitly coupling structure with appearance and weakening structural controllability under modality shifts. To address this, we propose \textbf{Path}ology \textbf{A}uto\textbf{R}gressive modeling (\textbf{PathAR}), a structure-first autoregressive synthesis framework that explicitly factorizes structure and appearance for modality-label-conditioned pathology generation.
PathAR employs a dual vector quantization (Dual-VQ) tokenizer to decompose samples into mask-grounded structure and appearance tokens, and an interleaved autoregressive (IAR) transformer with asymmetric attention visibility to enforce structure-to-appearance dependence. 
PathAR stabilizes morphology under heterogeneous modality-specific appearances and enables spatially aligned image--mask pair generation.
Extensive experiments show that PathAR improves structural consistency and modality fidelity over baselines, maintains sample diversity, supports downstream segmentation in data-scarce regimes, and demonstrates extensibility to finer-grained intra-modality organ-label variation.

\end{abstract}


\begin{figure}[t]
    \centering
    \includegraphics[width=\linewidth]{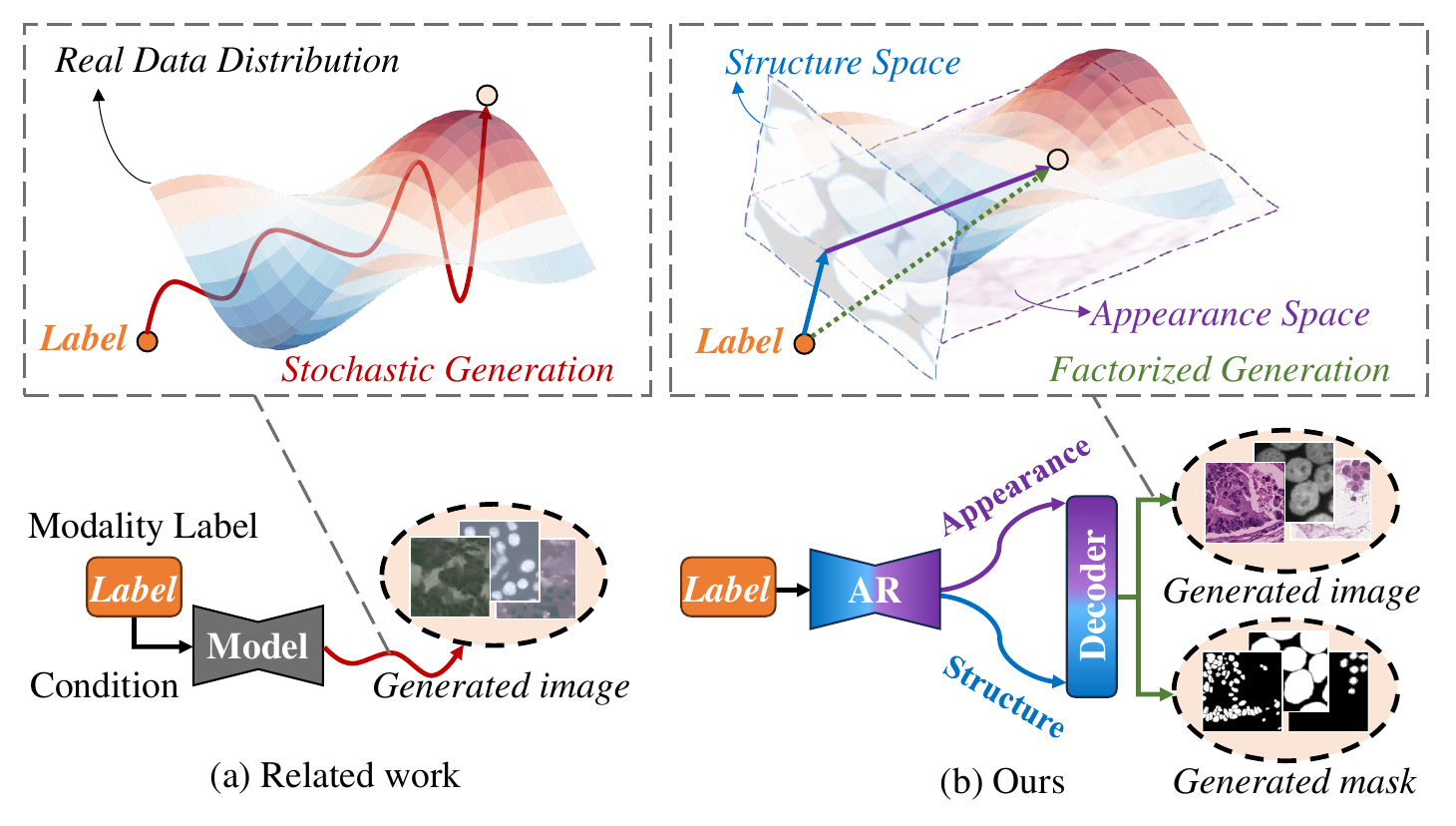}
    \caption{\textbf{Motivation of PathAR for pathology image generation conditioned on the modality label.} (a) Entangled stochastic generation can match modality-specific appearance but often disrupts morphology under multimodal heterogeneity. (b) PathAR uses factorized generation: it splits tokens into structure and appearance streams and enforces a structure-first autoregressive dependency to better maintain morphological consistency across modalities.}
    \label{fig:motivation}
\end{figure}
   
\section{Introduction}
\label{sec:introduction}
\IEEEPARstart{M}{ultimodal} pathology generation plays a key role in alleviating data scarcity under heterogeneous imaging domains \cite{pantanowitz2024synthetic, pathology2025survey}. 
While distinct modalities such as cytology, fluorescence, and histology exhibit significant visual heterogeneity due to divergent imaging mechanisms, their underlying morphological layout remains consistent across modalities \cite{ke2023clusterseg}. Thus, structural masks serve as a modality-agnostic spatial reference for morphology~\cite{Structure2016StainNorm, ragazzi2014fluorescence}. This motivates joint generation of a structural mask and a modality-specific image, directly supporting structure-sensitive tasks such as segmentation.

Multimodal generation requires balancing anatomical fidelity, which ensures structural integrity \cite{qian2024maskfactory}, with modality fidelity, which governs appearance distribution~\cite{cohen2018hallucinateMedical}. Neglecting anatomical fidelity in favor of appearance optimization introduces anatomically implausible artifacts~\cite{Hallucinations2021ImageRecon}, rendering the synthetic data unsuitable for structure-sensitive applications.

Empirical observations under multimodal heterogeneity often reveal a mismatch between appropriate modality appearance and morphologically coherent structure (Fig.~\ref{fig:motivation} (a)). 
It stems from the implicit entanglement of structure and appearance within a homogeneous latent space. Within such entangled representations, the modeling process is prone to texture bias~\cite{texturebias2018}, where strong appearance signals overshadow the subtle gradients required for structural consistency. Fig.~\ref{fig:motivation} (b) shows the motivation for a structure-to-appearance factorization. Enforced via an asymmetric attention-visibility constraint, this dependency mitigates imbalance between structure and appearance, stabilizing morphology while preserving modality-specific detail. 
This calls for a method capable of explicitly factorizing the modeling process into discrete, controllable streams.

Recent autoregressive modeling (AR) has driven remarkable progress in generative tasks, spanning from natural language processing \cite{gpt4, llama, llama2} to computer vision \cite{wei2023autoregressivetracking, xiong2024VARsurvey}. Visual AR approaches estimate the data distribution by factorizing it into a sequence of conditional probabilities, implemented via sequential next-token \cite{LLamaGen, wang2024emu3, wu2025janus} or next-scale prediction \cite{VAR, imagefolder, controlar} over a flattened stream of discrete indices. While effective for general generation, this formulation implicitly entangles geometry and texture, risking structural corruption under appearance shifts. Yet, the sequential nature of AR offers a unique opportunity for factorization: it allows us to impose a causal hierarchy by explicitly prioritizing structure before appearance.

To address this limitation, we propose PathAR, a structure-first autoregressive framework for modality-label-conditioned pathology image generation that factorizes structure and appearance. 
At the representation level, we design a dual vector quantization (Dual-VQ) tokenizer to construct two spatially aligned discrete streams: the structure stream is grounded by mask supervision during training, while the appearance stream captures modality-specific visual statistics, thereby reducing structure--appearance representation entanglement. 
At the generation level, we develop an interleaved autoregressive transformer to model the modality-conditioned joint prior under an asymmetric attention visibility constraint: structure tokens are predicted without attending to appearance history, and appearance is synthesized via conditional appearance rendering anchored to the predicted structure.
This explicit structure-first causality mitigates texture bias and stabilizes morphological consistency under multimodal heterogeneity. 
Thus, PathAR enables mask-free label-to-image generation, producing morphologically coherent and modality-faithful images alongside aligned auxiliary masks. 

In summary, our main contributions are as follows:

\begin{itemize}
\item We introduce PathAR, the first autoregressive framework for modality-label-conditioned multimodal pathology synthesis. PathAR jointly generates aligned image--mask pairs from modality labels alone at inference, enabling coherent morphology and modality-specific image synthesis.

\item We propose a Dual-VQ tokenizer to factorize samples into spatially aligned structure and appearance token streams, mitigating representation entanglement and enabling explicit structure-aware generation.

\item We design an interleaved autoregressive transformer with an asymmetric visibility constraint that enforces structure-first appearance rendering, establishing a directional dependency during generation.

\item Extensive experiments on multimodal pathology benchmarks demonstrate that PathAR improves structural fidelity and modality consistency over state-of-the-art baselines while maintaining sample diversity. The generated image--mask pairs further provide effective augmentation for downstream segmentation, with additional validation under finer intra-modality organ-label conditioning.
\end{itemize}

The remainder of this paper is organized as follows: Section~\ref{sec:relatedwork} reviews related work.
Section~\ref{sec:method} presents the proposed PathAR framework in detail. Section~\ref{sec:experiments} reports experimental comparisons, ablation studies, and model analyses. Section~\ref{sec:conclusion} concludes this paper.

\section{Related Work}
\label{sec:relatedwork}
\subsection{Pathology Image Generation}
Pathology image generation requires structural fidelity beyond global visual realism, as subtle structural artifacts can compromise clinical interpretation~\cite{singh2022nuclearMorphology}. To improve anatomical fidelity, prevailing methods typically rely on explicit spatial priors, using either external masks or source images as guidance~\cite{pathology2025survey}. This dependence restricts generation to conditional rendering and limits label-only generation of aligned image--mask pairs. (1) The first prominent paradigm focuses on structural control via \textbf{mask-conditioned generation}. Models such as PathDiff~\cite{pathdiff}, NASDM~\cite{nasdm}, and PathopixGAN~\cite{PathoGAN} formulate synthesis as mapping a given semantic layout to a textured image. While effective at boundary adherence, this paradigm treats structure as a prerequisite input rather than a generative target. As a result, these methods depend on masks and cannot generate diverse phenotypes when annotations are scarce.
(2) The second paradigm addresses \textbf{modality translation} across stains or protocols. Using source images as structural anchors, these methods preserve anatomical organization while transforming appearance~\cite{pati2024accelerating, latonen2024virtualStain}. Fundamentally, these approaches prioritize domain translation over generative sampling. Because they treat anatomical structure as an inference-time prerequisite, they cannot create paired data from scratch~\cite{de2021StainTransfer}. Consequently, label-conditional generation of aligned image--mask pairs remains underexplored. PathAR closes this gap by autoregressively generating morphology, enabling controllable generation under cross-modality appearance shifts.

\subsection{Visual Autoregressive Modeling} 
Discrete autoregressive modeling has emerged as a promising paradigm for visual synthesis by extending the sequential factorization of language models to images represented as tokens~\cite{xiong2024VARsurvey, chen2024nexttokensurvey}. 
This formulation maps images to sequences via vector-quantized (VQ) discretization~\cite{vqvae, VQGAN}, supporting masked prediction methods such as MaskGIT~\cite{maskgit} as well as autoregressive generation.
Given discrete tokens, AR uses strict causal dependencies to explicitly define information flow. 
Accordingly, visual AR adopts different serialization orders, including raster-scan next-token prediction~\cite{LLamaGen}, scale-wise progression in VAR~\cite{VAR}, and token folding in ImageFolder~\cite{imagefolder} to improve decoding efficiency.
However, despite these ordering choices, most prior methods still model images with a homogeneous token stream. This homogeneous representation implicitly entangles geometry and texture, rendering these methods ill-suited for multimodal tasks requiring precise structure. Medical image
translation studies similarly suggest that separating structure from intensity variation improves robustness to appearance shifts~\cite{zhang2024structureintensity}.
To address this, PathAR introduces a dual-stream factorization strategy, using interleaved attention to explicitly condition appearance synthesis on a generated structural scaffold.

\subsection{Structure-Guided and Controllable Generation}
Structure-guided controllable generation typically relies on explicit spatial conditions to dictate geometry during inference. 
GAN-based methods, including semantic-layout and edge-guided synthesis~\cite{SPADE, tan2021efficient, tang2023edge}, often utilize spatial layouts or boundary cues to enforce layout adherence.
For diffusion backbones such as DDPM~\cite{DDPM} and DiT~\cite{DiT}, structural control is commonly introduced by augmenting the denoiser with auxiliary conditioning networks. Specifically, ControlNet~\cite{ControlNet} and T2I-Adapter~\cite{t2i} inject provided structural hints, such as edges or masks, into intermediate denoising features. 
Similar strategies have been extended to AR models in ControlAR~\cite{controlar}.
A shared limitation of these paradigms is their operational reliance on external structure. 
They treat the mask as a condition rather than generating it as an explicit output, which does not directly support label-only synthesis of aligned image--mask pairs. While PathAR uses masks during training to learn structural priors, it treats the structure mask as a generative output rather than a prerequisite input. This design bridges the gap between label-only guidance and high-fidelity spatial synthesis during inference.

\section{Method}
\label{sec:method}
\subsection{Problem Formulation and Solution Overview}
\label{sec:problem}

\begin{figure*}[!t]
    \centering
    \includegraphics[width=0.97\textwidth]{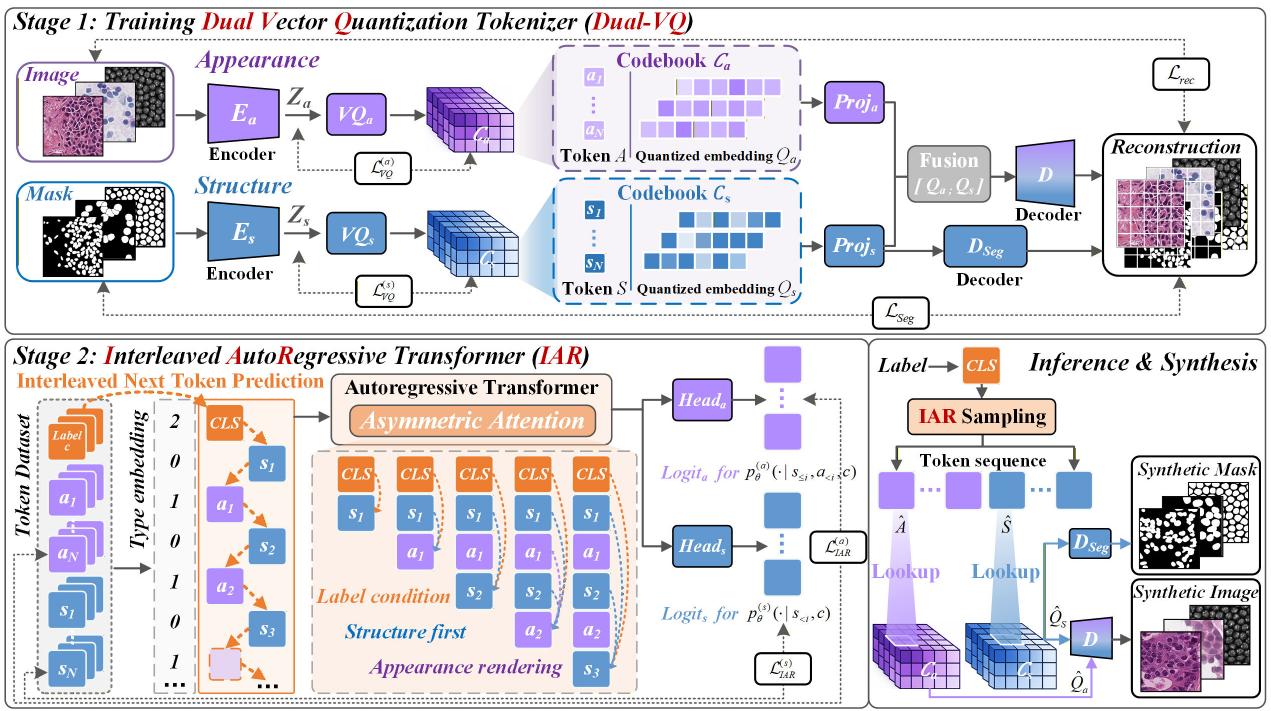}
    \caption{\textbf{Overview of PathAR.} \textbf{\textit{Stage 1}}: Dual-VQ factorizes pathological images into structure and appearance token streams, where the former is anchored by mask-guided reconstruction to yield discrete, semantically rich tokens. \textbf{\textit{Stage 2}}: The interleaved autoregressive transformer (IAR) models the modality-conditional joint token distribution via next-token prediction under asymmetric attention visibility, enforcing structure-to-appearance causality for appearance rendering.
    \textbf{\textit{Inference \& Synthesis}}: Given a modality label, IAR generates structure and appearance token sequences, which are decoded via codebook lookup into a synthetic image--mask pair.} 
    \label{fig:framework}
\end{figure*}

\textbf{Problem Formulation: }Given a modality label $c$ (e.g., cytology, fluorescence, histology), we aim to generate a modality-specific red-green-blue (RGB) pathology patch $x \in \mathbb{R}^{H \times W \times 3}$, where $H$ and $W$ denote the patch height and width, by modeling $p(x \mid c)$.
In multimodal settings, appearance varies across acquisition mechanisms, while morphological structure remains stable under the same annotation semantics.
To make morphology explicit, we additionally model paired generation with an aligned structural mask $m \in \{0,1\}^{H \times W}$:
\begin{equation}
\label{eq:pair_goal}
p(x,m\mid c).
\end{equation}
With a tokenizer downsampling factor $r$, the latent grid is $h\times w$ with $h=H/r$, $w=W/r$, and $N=hw$ tokens.
An overview of the proposed framework is shown in Fig.~\ref{fig:framework}.

\paragraph{From the chain rule to paired synthesis}
By the probabilistic chain rule, the joint distribution in Eq.~\eqref{eq:pair_goal} can be factorized as:
\begin{equation}
\label{eq:chain_pair}
p(x,m\mid c)=p(m\mid c)\,p(x\mid m,c),
\end{equation}
which makes structure explicit via $p(m\mid c)$.

\paragraph{Discrete latent factorization}
We introduce two discrete latent variables:
\emph{structure tokens} $S=(s_1,\dots,s_N)$ and \emph{appearance tokens} $A=(a_1,\dots,a_N)$ on the $h\times w$ latent grid.
Let $p_{\psi}(m\mid S)$ denote the mask likelihood parameterized by a structure decoder and $p_{\phi}(x\mid S,A)$ the image likelihood parameterized by an image decoder.
Paired generation is obtained by integrating out the discrete latents $(S,A)$:
\begin{equation}
\label{eq:latent_factorization}
\resizebox{0.7\linewidth}{!}{
$\begin{aligned}
p_{\Theta}(x,m\mid c)
&=\sum_{S,A}\underbrace{p_{\theta}(S\mid c)}_{\text{structure prior}}
\cdot \underbrace{p_{\theta}(A\mid S,c)}_{\text{appearance prior}} \\
&\quad\cdot \underbrace{p_{\psi}(m\mid S)}_{\text{structure decoding}}
\cdot \underbrace{p_{\phi}(x\mid S,A)}_{\text{image decoding}}.
\end{aligned}$}
\end{equation}
Here, $\theta$ parameterizes the token priors (implemented by our interleaved autoregressive transformer), and $\psi$ and $\phi$ parameterize the mask decoder $D_{\mathrm{seg}}$ and the image decoder $D$, respectively.
We denote overall parameters by $\Theta=\{\theta,\phi,\psi\}$.

\paragraph{Variational Training Objective}
Given training tuples $(x,m,c)$, we introduce an approximate posterior produced by the Dual-VQ tokenizer,
$q_{\varphi}(S,A\mid x,m)$ with tokenizer parameters $\varphi$.
Using the factorization
$p_{\theta}(S,A\mid c)=p_{\theta}(S\mid c)\,p_{\theta}(A\mid S,c)$, the evidence lower bound (ELBO) is:
\begin{equation}
\label{eq:elbo}
\resizebox{0.9\linewidth}{!}{$
\begin{aligned}
& \log p_{\Theta}(x,m\mid c)
\ge~\mathbb{E}_{q_{\varphi}(S,A\mid x,m)}\Big[
\log p_{\phi}(x\mid S,A) \\
& + \log p_{\psi}(m\mid S) 
\Big] 
-\mathrm{KL}\!\left(q_{\varphi}(S,A\mid x,m)\,\|\,p_{\theta}(S,A\mid c)\right).
\end{aligned}
$}
\end{equation}
With a near-deterministic $q_{\varphi}$, Eq.~\eqref{eq:elbo} motivates a two-stage protocol:
(i) optimize tokenizer and decoders $\{\varphi,\phi,\psi\}$ via the reconstruction and segmentation terms (Eq.~\eqref{eq:tokenizer_loss});
(ii) fix $\{\varphi,\phi,\psi\}$ and minimize KL divergence, which reduces to fitting the prior $p_{\theta}(S,A\mid c)$ via maximum likelihood (Eq.~\eqref{eq:ar_loss}).

\subsection{Dual-VQ Tokenizer}
\label{sec:dvq}
Given a training pair $(x,m)$, the tokenizer produces two spatially aligned token streams: a \emph{structure} stream $S$ anchored by the mask $m$, and an \emph{appearance} stream $A$ that encodes modality-specific residuals complementary to $S$.

\subsubsection{Asymmetric encoders and discretization}
Two encoders with asymmetric information access map the pair $(x,m)$ to latent grids.
Structure encoder $E_s$ maps $m$:
\begin{equation}
Z_s = E_s(m)\in\mathbb{R}^{h\times w\times d},
\end{equation}
and the appearance encoder $E_a$ maps the image $x$:
\begin{equation}
Z_a = E_a(x)\in\mathbb{R}^{h\times w\times d},
\end{equation}
where $(h,w)$ is the tokenizer latent grid resolution, $N=hw$ is the number of spatial tokens, and $d$ is the latent embedding dimension.
Let $i\in\{1,\dots,N\}$ index spatial locations on the $h\times w$ grid, and denote
$z^{(s)}_i = Z_s[i]\in\mathbb{R}^d$ and $z^{(a)}_i = Z_a[i]\in\mathbb{R}^d$.

Two codebooks are learned,
$\mathcal{C}_s=\{e^{(s)}_1,\dots,e^{(s)}_{K_s}\}$ and
$\mathcal{C}_a=\{e^{(a)}_1,\dots,e^{(a)}_{K_a}\}$ with $e^{(s)}_k,e^{(a)}_k\in\mathbb{R}^d$. Here $K_s$ and $K_a$ denote the structure and appearance codebook sizes, respectively.
Quantization applies nearest-neighbor assignment independently at each location:
\begin{equation}
\resizebox{0.9\linewidth}{!}{$
\begin{aligned}
s_i &= \arg\min_{k\in\{1,\dots,K_s\}} \big\|z^{(s)}_i - e^{(s)}_k\big\|_2, 
&\ q^{(s)}_i &= e^{(s)}_{s_i},\\
a_i &= \arg\min_{k\in\{1,\dots,K_a\}} \big\|z^{(a)}_i - e^{(a)}_k\big\|_2,
&\  q^{(a)}_i &= e^{(a)}_{a_i}.
\end{aligned}$}
\end{equation}
The resulting index sequences are $S=(s_1,\dots,s_N)$ and $A=(a_1,\dots,a_N)$, with corresponding quantized embeddings
$Q_s=(q^{(s)}_1,\dots,q^{(s)}_N)$ and $Q_a=(q^{(a)}_1,\dots,q^{(a)}_N)$.
This encoder--quantizer pipeline defines an approximate posterior $q_{\varphi}(S, A \mid x,m)$, which is typically near-deterministic in practice; $\varphi$ denotes tokenizer parameters.

\subsubsection{Decoders and factorized supervision}
\label{sec:decoding_dynamics}
To ground structure tokens and enforce complementary appearance encoding, we use two decoding heads on quantized embeddings: segmentation decoder $D_{\mathrm{seg}}$ predicts masks from structure codes only, while image decoder $D$ outputs RGB images from fused structure--appearance codes:
\begin{equation}
\label{eq:decoder}
\hat{m}=D_{\mathrm{seg}}(Q_s;\psi),\qquad 
\hat{x}=D\!\left([Q_s;\,Q_a];\phi\right).
\end{equation}
Here $[\cdot\,;\cdot]$ denotes a learnable feature fusion operation that projects $Q_s$ and $Q_a$ into a common decoder feature space and fuses them through scaled structural conditioning, channel-wise concatenation, and pointwise convolution. This design uses $Q_s$ as a morphology-grounded structural condition and $Q_a$ as a complementary source of modality-specific appearance information.

Following Eq.~\eqref{eq:elbo}, we optimize the likelihood terms $\log p_{\phi}(x\mid S,A)$ and $\log p_{\psi}(m\mid S)$ via reconstruction and segmentation losses that serve as negative log-likelihood surrogates, and add standard VQ regularization, yielding Eq.~\eqref{eq:tokenizer_loss}:
\begin{equation}
\label{eq:tokenizer_loss}
\resizebox{0.7\linewidth}{!}{
$
\begin{aligned}
\mathcal{L}_{\mathrm{DVQ}}
&=
\mathcal{L}_{\mathrm{rec}}(x,\hat{x})
+\mathcal{L}^{(s)}_{\mathrm{VQ}}(Z_s,Q_s;\mathcal{C}_s) \\
&+\mathcal{L}^{(a)}_{\mathrm{VQ}}(Z_a,Q_a;\mathcal{C}_a)
+\mathcal{L}_{\mathrm{seg}}(m,\hat{m}).
\end{aligned}
$
}
\end{equation}
\noindent
Here $\mathcal{L}_{\mathrm{rec}}$ denotes the image reconstruction loss between $x$ and $\hat{x}$, and $\mathcal{L}_{\mathrm{seg}}$ denotes the mask prediction loss between $m$ and $\hat{m}$. Each $\mathcal{L}_{\mathrm{VQ}}$ term follows the standard stop-gradient VQ regularization~\cite{vqvae}, and is applied independently to the corresponding structure or appearance codebook.
We use an unweighted sum of loss terms.
This interface encourages a functional separation: $\mathcal{L}_{\mathrm{seg}}$ makes $Q_s$ predictive of $m$, while $\mathcal{L}_{\mathrm{rec}}$ encourages $Q_a$ to encode appearance information complementary to $Q_s$, without explicit orthogonality constraints.

\subsection{Interleaved Autoregressive Transformer}
\label{sec:iar}
The token priors are parameterized by a single interleaved autoregressive transformer that generates structure and appearance tokens within one sequence under an asymmetric attention constraint. Concretely, two token streams $(S,A)$ are interleaved by pairing tokens at each spatial location:
\begin{equation}
\label{eq:interleaved_seq}
U=\big[\mathrm{CLS}(c),\, s_1,a_1,\, s_2,a_2,\,\dots,\, s_N,a_N\big],
\end{equation}
where $c$ is the modality label condition and is injected only via the CLS prefix token $\mathrm{CLS}(c)$.
Let $u_t$ be the $t$-th token in $U$, so that $u_{2i}=s_i$ and $u_{2i+1}=a_i$. The joint prior follows the standard autoregressive factorization:
\begin{equation}
\label{eq:ar_factor}
\resizebox{0.88\linewidth}{!}{
$
p_{\theta}(U\mid c)=\prod_{t=2}^{2N+1} p_{\theta}(u_t\mid u_{<t},c),
\quad \text{with } u_1=\mathrm{CLS}(c).
$}
\end{equation}

\paragraph{Explicit dependency via asymmetric attention}
The interleaved generator encodes a structure-first dependency via asymmetric attention visibility. When predicting a structure token $s_i$, the attention mask permits access only to the modality label and previous structure tokens $s_{<i}$, while blocking all appearance tokens $a_{<i}$ from every self-attention layer, thereby learning a structural prior that is not conditioned on
generated appearance history. In contrast, when predicting an appearance token $a_i$, the model can attend to $s_{\le i}$ and
$a_{<i}$, enabling structure-aware appearance rendering:
\begin{align}
\label{eq:cond_s}
& p_{\theta}(s_i \mid u_{<2i},c) = p_{\theta}(s_i \mid s_{<i},c),\\
\label{eq:cond_a}
& p_{\theta}(a_i \mid u_{<2i+1},c) = p_{\theta}(a_i \mid s_{\le i},a_{<i},c).
\end{align}
This asymmetric visibility implements the intended directional dependency from structure to appearance.
2D Rotary Positional Embedding (2D-RoPE)~\cite{ROPE} is applied on the $h\times w$ grid, where $(\rho_i,\kappa_i)$ denotes the row and column coordinates of location $i$, with paired position
reuse $\mathrm{Pos}(s_i)=\mathrm{Pos}(a_i)=(\rho_i,\kappa_i)$, which ensures pixel-aligned structure--appearance correspondence when predicting $a_i$.
With teacher forcing, two heads parameterize structure and appearance conditions, $p_{\theta}^{(s)}$ and $p_{\theta}^{(a)}$. Under this visibility constraint, the negative log-likelihood (NLL) becomes:
\begin{equation}
\label{eq:nll_decomp}
\resizebox{0.88\linewidth}{!}{
$\begin{aligned}
-\log p_{\theta}(U\mid c)
&= \sum_{t=2}^{2N+1} -\log p_{\theta}\!\left(u_t \mid u_{<t},c\right) \\
&= \sum_{i=1}^{N}\Big(
-\log p^{(s)}_{\theta}(s_i\mid s_{<i},c)
-\log p^{(a)}_{\theta}(a_i\mid s_{\le i},a_{<i},c)
\Big).
\end{aligned}$}
\end{equation}
The conditional forms follow from the visibility constraint (Eq.~\eqref{eq:cond_s}--\eqref{eq:cond_a}).
Equivalently, minimizing the sequence NLL corresponds to the cross-entropy objective:
\begin{equation}
\label{eq:ar_loss}
\resizebox{0.89\linewidth}{!}{
$\begin{aligned}
\mathcal{L}_{\mathrm{IAR}}
&= \underbrace{\sum_{i=1}^{N} \mathrm{CE}\!\left(p_{\theta}^{(s)}(\cdot \mid s_{<i},c),\, s_i\right)}_{\mathcal{L}_{\mathrm{IAR}}^{(s)}}
+ \underbrace{\sum_{i=1}^{N} \mathrm{CE}\!\left(p_{\theta}^{(a)}(\cdot \mid s_{\le i},a_{<i},c),\, a_i\right)}_{\mathcal{L}_{\mathrm{IAR}}^{(a)}}.
\end{aligned}$}
\end{equation}
Minimizing $\mathcal{L}_{\mathrm{IAR}}$ fits $p_{\theta}(S,A\mid c)$ by maximum likelihood, corresponding to the KL term in Eq.~\eqref{eq:elbo}.

\subsection{Inference and Sampling}
\label{sec:inference}
During inference, we sample from the interleaved autoregressive prior $p_{\theta}(U\mid c)$ in Eq.~\eqref{eq:ar_factor}.
Given the label token $\mathrm{CLS}(c)$, tokens are generated sequentially to obtain an interleaved sample
$\hat{U}$, which is split into paired streams $(\hat{S},\hat{A})$.
The same attention constraint used during training (Eq.~\eqref{eq:cond_s}--\eqref{eq:cond_a}) is preserved during sampling,
thereby enforcing structure-first generation and structure-conditioned rendering.
The sampled indices are mapped to quantized embeddings by codebook lookup:
\begin{equation}
\resizebox{0.5\linewidth}{!}{
$\hat{Q}_s=\mathcal{C}_s[\hat{S}],\qquad \hat{Q}_a=\mathcal{C}_a[\hat{A}].$
}
\end{equation}
The synthetic image--mask pairs $(\tilde{x},\tilde{m})$ are then obtained by deterministic decoding using the decoders in Eq.~\eqref{eq:decoder}.
During sampling, standard classifier-free guidance (CFG)~\cite{CFG} is optionally applied to the token logits to improve label consistency, without altering the interleaved order or the asymmetric visibility constraint.

\section{Experiments}
\label{sec:experiments}

\subsection{Experimental Setup}
\label{sec:experimental_settings}

\paragraph{Task}
We define two inference tasks according to the available inference-time inputs.

(1) \textbf{(L+M)2I:} Synthesis conditioned on the modality label and ground-truth mask to generate images.

(2) \textbf{L2I:} Synthesis conditioned solely on the modality label to generate images. PathAR follows this label-only setting while additionally inferring an auxiliary mask.

\paragraph{Dataset}
We use three public multimodal pathology subsets~\cite{ke2023clusterseg}: cytology (487 $224\times224$ images), fluorescence (524 $512\times512$ images), and histology (462 $512\times512$ images), with a $9{:}1$ train/validation split.
All samples are paired with instance-level nuclear annotations, from which binary segmentation masks are derived.
We additionally evaluate PathAR on PanNuke~\cite{gamper2020pannuke} for organ-label-conditioned histopathology generation, following the official $2{:}1$ train/validation split.
PanNuke comprises 7,901 $256\times256$ histopathology images annotated with 19 organ labels. Unlike the primary multimodal benchmark, where the condition label $c$ indicates the imaging modality, $c$ in PanNuke specifies the organ label under a single histopathological modality.

\paragraph{Evaluation metrics}
Generation quality is evaluated using FID~\cite{fid} and KID~\cite{kid} computed on DINOv2~\cite{dinov2} and UNI~\cite{uni} feature embeddings.
We denote these variants as FID-DINOv2, KID-UNI, and KID-DINOv2, respectively.
KID provides an unbiased estimate of distributional discrepancy, which is useful for small-scale medical datasets.
For readability, all FID and KID values in the tables and text are reported in units of $10^{-3}$ unless otherwise specified.
For appearance fidelity, we report Style Score, defined as the modality classification accuracy of a classifier trained on real samples.
For structural fidelity, we report Dice~\cite{dice1945measures} from downstream segmentation models trained on generated image--mask pairs and evaluated on real validation data.
For sample diversity, we report LPIPS~\cite{zhang2018unreasonable} and Recall~\cite{kynkaanniemi2019improved}.

\paragraph{Implementation details}
We optimize PathAR using AdamW with $\beta_1=0.9$, $\beta_2=0.95$, a weight decay of $0.05$, and a learning rate of $1\times10^{-4}$. The Dual-VQ tokenizer uses a downsampling factor of $r=16$, codebook sizes of $K_s=K_a=1024$, and an embedding dimension of $d=16$. The IAR transformer follows a GPT-style architecture with 8 layers, 8 attention heads, and a hidden dimension of 512, and is trained for 300 epochs with a batch size of 32. All experiments are conducted on two NVIDIA RTX A6000 GPUs.

\begin{table*}[!t]
    \centering
    \caption{Quantitative comparison. We compare PathAR with 16 baselines, including (L+M)2I mask-conditioned methods (top rows) and L2I label-only methods (bottom rows).
    Despite requiring no mask input at inference, PathAR achieves competitive or superior image fidelity against mask-conditioned (L+M)2I approaches while generating coherent morphology.
    Model architecture: DM = diffusion model, AR = autoregressive model, GAN = generative adversarial network; * indicates a pathology-specific model. Modality: Cyto., Fluo., His. denote cytology, fluorescence, and histology, respectively. The best results are shown in \textbf{bold}. The second-best results are \underline{underlined}.}
    \resizebox{\textwidth}{!}{
    \begin{tabular}{c|l|c|cccc|cccc|cccc}
    \toprule
         \multirow{2}{*}{\textbf{Task}}& \multirow{2}{*}{\textbf{Method}} & \multirow{2}{*}{\textbf{Architecture}}&  \multicolumn{4}{c|}{\textbf{FID-DINOv2} $\downarrow$}&  \multicolumn{4}{c|}{\textbf{KID-UNI} $\downarrow$}&  \multicolumn{4}{c}{\textbf{KID-DINOv2} $\downarrow$}\\
         \cline{4-15}
         ~& ~ & ~ & Cyto. & Fluo. & His. & Avg. & Cyto. & Fluo. & His. & Avg. & Cyto. & Fluo. & His. & Avg. \\
    \midrule
               \multirow{8}{*}{\textbf{(L+M)2I}}& SPADE~\cite{SPADE} & GAN &  230.301&  168.268&  334.972&  244.514&  1.848&  0.335&  1.085&  1.089&  1.149&  \underline{0.228}&  1.205&  0.860
\\ 
         ~ & PathopixGAN~\cite{PathoGAN} & GAN* & 131.039 &  \textbf{95.620}&  378.107&  201.589&  1.002&  \textbf{0.105}&  0.846&  \underline{0.651}&  1.091&  \textbf{0.079}&  1.250&  0.807
\\ 
         ~ & ControlNet~\cite{ControlNet} & DM &  142.260&  245.617&  240.217&  209.365&  1.608&  0.330&  \underline{0.506}&  0.814&  0.458&  0.384&  \textbf{0.429}&  \underline{0.423}
\\ 
         ~ & DiffInfinite~\cite{diffinfinite} & DM* &  686.285&  669.875&  709.358&  688.506&  2.630&  1.267&  2.445&  2.114&  3.114&  1.949&  2.256&  2.440
\\
         ~ & NASDM~\cite{nasdm} & DM* &  473.765&  291.547&  373.326&  379.546&  1.908&  0.594&  1.343&  1.282&  2.483&  0.618&  1.891&  1.664
\\ 
         ~ & T2I-Adapter~\cite{t2i} & DM &  241.484&  357.923&  390.930&  330.112&  1.404&  0.469&  1.251&  1.041&  1.289&  0.921&  1.463&  1.224
\\ 
         ~ & PathDiff~\cite{pathdiff} & DM* &  \textbf{91.378}&  244.984&  \underline{100.422}&  \underline{145.595}&  \textbf{0.931}&  0.902&  0.511&  0.781&  \textbf{0.443}&  0.758&  0.459&  0.553
\\ 
         ~ & ControlAR~\cite{controlar} & AR &  1062.378&  1091.804&  973.665&  1042.616&  2.261&  1.381&  1.777&  1.806&  2.726&  2.889&  2.457&  2.691
\\
    \midrule
         \multirow{10}{*}{\textbf{L2I}}& ADA~\cite{ADA} & GAN &  473.227&  390.668&  263.456&  375.784&  1.525&  0.809&  0.771&  1.035&  1.666&  1.485&  0.828&  1.327
\\ 
         ~ & VQ-GAN~\cite{VQGAN} & GAN &  257.017&  595.675&  210.404&  354.365&  1.114&  0.568&  0.447&  0.710&  2.644&  1.665&  1.059&  1.790
\\ 
         ~ & DDPM~\cite{DDPM} & DM &  581.958&  384.279&  428.839&  465.025&  1.657&  1.365&  1.254&  1.425&  2.929&  1.432&  1.654&  2.005
\\ 
         ~ & Stable Diffusion~\cite{StableDiffusion} & DM &  358.178&  335.204&  334.372&  342.585&  0.997& 0.233&  0.753&  0.661&  0.903&  0.869&  1.261&  1.011
\\ 
         ~ & DiT~\cite{DiT} & DM &  162.715&  \underline{136.502}&  185.013&  161.410&  0.960&  1.163&  0.665&  0.929&  0.948&  0.385&  0.800&  0.711
\\ 
         ~ & LlamaGen~\cite{LLamaGen} & AR &  231.841&  463.199&  216.107&  303.716&  \underline{0.946}&  0.347&  0.858&  0.717&  0.576&  0.610&  1.001&  0.729
\\
         ~ & VAR~\cite{VAR} & AR &  424.466&  399.566&  495.556&  439.863&  1.590&  0.415&  1.057&  1.021&  1.072&  1.318&  1.145&  1.179
\\ 
         ~ & ImageFolder~\cite{imagefolder} & AR &  452.915&  483.312&  356.640&  430.956&  2.721&  0.527&  1.188&  1.478&  3.069&  0.647&  1.667&  1.795
\\
        \cline{2-15}
        \rowcolor{gray!20}
       \cellcolor{white}~ & \cellcolor{gray!20}\textbf{PathAR (ours)} & AR* & 
       \underline{128.283}&  167.541&  \textbf{95.834}&  \textbf{130.552}&  1.049& \underline{0.186}&  \textbf{0.495}&  \textbf{0.576}&  \underline{0.459}&  0.235&  \underline{0.454}&  \textbf{0.382}
\\
    \bottomrule
    \end{tabular}}
    \label{tab:1}
\end{table*}

\paragraph{Baselines}
We benchmark PathAR against 16 representative generative baselines spanning GAN, diffusion, and autoregressive paradigms, covering both general-vision and pathology-specific models. (1) GAN: SPADE~\cite{SPADE}, StyleGAN-ADA~\cite{ADA}, VQGAN~\cite{VQGAN}.
(2) Diffusion: DDPM~\cite{DDPM}, Stable Diffusion~\cite{StableDiffusion}, DiT~\cite{DiT}, ControlNet~\cite{ControlNet}, T2I-Adapter~\cite{t2i}.
(3) AR: LlamaGen~\cite{LLamaGen}, VAR~\cite{VAR}, ImageFolder~\cite{imagefolder}, ControlAR~\cite{controlar}.
(4) Pathology-specific: PathopixGAN~\cite{PathoGAN}, PathDiff~\cite{pathdiff}, DiffInfinite~\cite{diffinfinite}, NASDM~\cite{nasdm}.

All baselines are implemented with their official open-source settings and recommended configurations.
Except for DDPM, all baselines are trained under the same multimodal joint setting as PathAR, where a single model is learned from the pooled cytology, fluorescence, and histology training data and conditioned on the modality label when supported.
Since the evaluated DDPM implementation is unconditional, we train separate DDPM models for each modality.
For tokenizer-based AR baselines, we retain each method's native tokenizer--generator design: VAR and ImageFolder use the released pretrained tokenizers, while LlamaGen and ControlAR retrain their tokenizers on the same pathology split using the official tokenizer training protocols.

\begin{figure*}[!t]
    \centering
    \includegraphics[width=\textwidth]{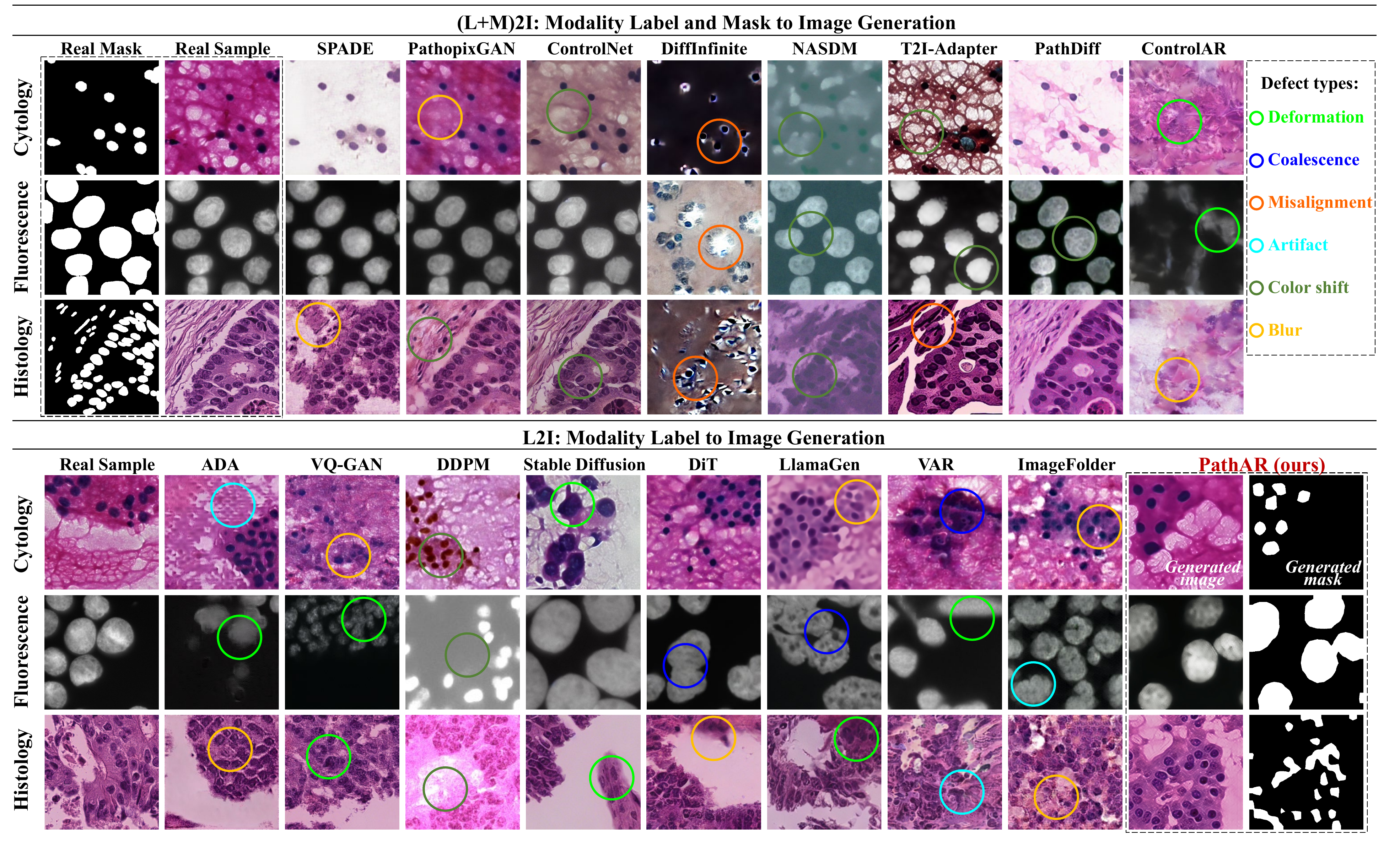}
    \caption{\textbf{Qualitative comparison of multimodal pathology generation at $256\times256$.} 
\textbf{Top:} (L+M)2I. 
\textbf{Bottom:} L2I. Generally, (L+M)2I stabilizes structure but may distort appearance across modalities, whereas L2I preserves modality appearance but often degrades structure without spatial anchors. PathAR (ours) maintains both structure and appearance in L2I and additionally generates an auxiliary mask.
Representative defects are circled: structural (\textcolor[RGB]{0,180,0}{deformation}, \textcolor{blue}{coalescence}, \textcolor[RGB]{255,102,0}{misalignment}) and textural (\textcolor[RGB]{0,180,180}{artifacts}, \textcolor[RGB]{84,130,53}{color shift}, \textcolor[RGB]{255,192,0}{blur}).}
    \label{fig:result_qualitative}
\end{figure*}

\subsection{Comparison with State-of-the-Art Methods}
\label{sec:sota_comparison}
The evaluation of pathology generation extends beyond distributional image quality, as generated samples are expected to preserve modality-specific appearance, maintain biologically plausible structures, and retain sufficient diversity.
Accordingly, we compare PathAR with representative methods across complementary aspects, including distributional fidelity,
qualitative morphology and texture, appearance consistency, structure-aware downstream segmentation, and generated sample
diversity.

\begin{table}[!t]
    \centering
    \caption{Evaluation of appearance fidelity via style classification. We report Style Score as the accuracy of modality classification of generated samples. Higher values indicate stronger consistency with the conditioning modality.}
    \resizebox{\linewidth}{!}{
    \begin{tabular}{c|l|cccc}
    \toprule
      \multirow{2}{*}{\textbf{Task}} & \multirow{2}{*}{\textbf{Method}} & \multicolumn{4}{c}{\textbf{Style Score} $\uparrow$} \\
       \cline{3-6}
      ~ & ~ & Cyto. & Fluo. & His. & Avg.\\
    \midrule
       \multirow{8}{*}{\textbf{(L+M)2I}} & SPADE~\cite{SPADE}   &  0.915&  0.962& 0.936 & 0.937\\
       ~ & PathopixGAN~\cite{PathoGAN} & 0.776 & \textbf{1.000} & \textbf{0.998} & 0.925\\
       ~ & ControlNet~\cite{ControlNet}  & 0.425 & 0.992 & \underline{0.992} & 0.803\\
       ~ & DiffInfinite~\cite{diffinfinite} & 0.250 & 0.225 & 0.590 & 0.355\\
       ~ & NASDM~\cite{nasdm}   & 0.052 & 0.893 & 0.281 & 0.409\\
       ~ & T2I-Adapter~\cite{t2i} & 0.675 & 0.865 & 0.760 & 0.767\\
       ~ & PathDiff~\cite{pathdiff} & \textbf{0.995} & 0.998 & 0.841 & 0.945\\
       ~ & ControlAR~\cite{controlar} & 0.374 & 0.694 & 0.638 & 0.568\\
    \midrule
      \multirow{9}{*}{\textbf{L2I}}
      & ADA~\cite{ADA} & 0.914 & 0.964 & 0.947 & 0.941\\
      ~ & VQ-GAN~\cite{VQGAN} & 0.866 & 0.630 & 0.970 & 0.822\\
      ~ & DDPM~\cite{DDPM} & 0.405 & 0.808 & 0.309 & 0.507\\
      ~ & Stable Diffusion~\cite{StableDiffusion} & 0.340 & 0.716 & 0.700 & 0.585\\
      ~ & DiT~\cite{DiT}  & 0.902 & \underline{0.999} & 0.958 & \underline{0.953}\\
      ~ & LlamaGen~\cite{LLamaGen}  & 0.253 & 0.594 & 0.667 & 0.505\\
      ~ & VAR~\cite{VAR}  & 0.882 & 0.977 & 0.968 & 0.942\\
      ~ & ImageFolder~\cite{imagefolder}  & 0.901 & 0.995 & 0.963 & \underline{0.953}\\
      \cline{2-6}
    \rowcolor{gray!20}
       \cellcolor{white}~ & \textbf{PathAR (ours)}  & \underline{0.944} & \textbf{1.000} & 0.968 & \textbf{0.970}\\
    \bottomrule
    \end{tabular}}
    \label{tab:result_classification}
\end{table}

\paragraph{Quantitative comparison}
Table~\ref{tab:1} reports quantitative comparisons on generated multimodal images across cytology, fluorescence, and histology, computed using 500 patches ($256\times256$) per modality.
PathAR achieves the best average FID-DINOv2 of 130.552, improving over the strongest baseline (PathDiff) by 15.043.
PathAR also consistently leads in KID-UNI and KID-DINOv2, with average scores of 0.576 and 0.382, indicating robust alignment across domain-specific and general features.
The improvement is particularly evident in
histology, where structural and textural
variations are more pronounced, whereas the narrower gaps in fluorescence are consistent with its comparatively constrained appearance distribution.
PathAR matches or outperforms mask-conditioned baselines without using mask input at inference, indicating structural layout can be learned as an explicit generative target rather than a mandatory condition.

\paragraph{Qualitative comparison}
Figure~\ref{fig:result_qualitative} qualitatively compares multimodal pathology image generation at $256 \times 256$ under the (L+M)2I and L2I settings.
(L+M)2I baselines are structurally stable due to explicit mask constraints, but often show appearance artifacts (e.g., blur or color shift) since semantic layout alone does not specify modality-specific appearance.
While L2I baselines effectively capture modality-level appearance and color statistics, they suffer from structural degradation due to the absence of spatial anchors, manifesting as nuclear coalescence, boundary degradation, and local shape deformation.
Pathology-specific generators (e.g., PathopixGAN and PathDiff) remain competitive on histology, but their generation quality becomes less stable under multimodal appearance heterogeneity, suggesting that pathology-specific designs alone are insufficient to robustly model diverse modality-specific appearances.
Bridging this gap, PathAR achieves stable morphology and fine-grained textural fidelity within the L2I setting.
This is realized through a structure-first mechanism, where latent structure tokens serve as an internal anchor to guide the autoregressive generation of appearance.
Furthermore, PathAR uniquely produces pixel-aligned masks alongside images, enabling direct verification of structural fidelity.

\begin{table}[!t]
    \centering
    \caption{Evaluation of structural fidelity via downstream segmentation. We report Dice on real validation data using nnU-Net trained exclusively on synthetic pairs ($N=100$). Higher scores indicate better anatomical coherence and image--mask alignment. The best results are shown in \textbf{bold}. The second-best results are \underline{underlined}.}
    \resizebox{\linewidth}{!}{
    \begin{tabular}{l|cccc}
    \toprule
       \multirow{2}{*}{\textbf{Method}} & \multicolumn{4}{c}{\textbf{Downstream Segmentation} $\uparrow$}\\
       \cline{2-5}
       ~  & Cyto. & Fluo. & His. & Avg.\\
    \midrule
       SPADE~\cite{SPADE}   & 0.905 & 0.950 & 0.797& 0.884
\\
       PathopixGAN~\cite{PathoGAN}  &0.899 & 0.950& 0.822& 0.890
\\
       ControlNet~\cite{ControlNet}  & 0.901& \textbf{0.955}& \underline{0.837}& 0.898
\\
       DiffInfinite~\cite{diffinfinite}  &0.647 & 0.836& 0.597& 0.693
\\
       NASDM~\cite{nasdm}  &0.897 &0.932 &0.729& 0.853
\\
       T2I-Adapter~\cite{t2i}  &0.873 &0.942&0.803 & 0.873
\\
       PathDiff~\cite{pathdiff}   & \textbf{0.915}&0.929 & 0.830& \underline{0.891}
\\
       ControlAR~\cite{controlar} & 0.773 & 0.541 & 0.595 & 0.636
\\
    \midrule
    \rowcolor{gray!20}
      \textbf{PathAR (ours)}  & \underline{0.906} & \underline{0.954} & \textbf{0.909} & \textbf{0.923}\\
    \bottomrule
    \end{tabular}}
    \label{tab:result_segmentation}
\end{table}

\paragraph{Appearance evaluation}
To assess modality-consistent appearance independently of structure, we adopt a proxy evaluation based on modality classification.
An EfficientNet-B0~\cite{efficientnet} classifier is trained on real samples to distinguish cytology, fluorescence, and histology, achieving $99.8\%$ validation accuracy.
The Style Score is defined as the classification accuracy of this fixed classifier on generated images, measuring the proportion of samples whose predicted modality matches the conditioning label.
As shown in Table~\ref{tab:result_classification}, PathAR achieves the highest average Style Score of 0.970, demonstrating stronger modality-specific appearance fidelity across the three modalities.
Fluorescence shows consistently high scores with small inter-method gaps, likely due to its distinctive photometric patterns.
Nevertheless, PathAR remains the best overall, indicating that it more effectively preserves the domain-specific texture and style statistics needed for reliable modality recognition.

\paragraph{Structure evaluation}
To quantify structural fidelity and practical utility, we evaluate whether generated data can supervise a segmentation model that generalizes to real validation data.
This downstream task requires paired images and masks; therefore, we compare the mask-conditioned baselines against PathAR, which provides an inferred mask via structure decoding in the L2I setting.
For each method and modality, we generate $100$ paired samples to train an nnU-Net~\cite{nnunet} under the same protocol, and report Dice~\cite{dice1945measures} on the real validation set.
Higher Dice indicates that the synthetic pairs encode consistent and transferable structural cues rather than merely plausible appearances.
As shown in Table~\ref{tab:result_segmentation}, PathAR achieves the best average Dice of 0.923, exceeding the second-best result of 0.898. The improvement is most pronounced on histology, where PathAR obtains a Dice of 0.909 compared with the second-best score of 0.837.
These results demonstrate that the decoded structure tokens provide segmentation-relevant morphology that effectively supervises downstream models, bridging autonomous generation and structural utility.

\begin{table*}[!t]
    \centering
    \caption{Impact of synthetic data scaling on downstream segmentation. Dice scores are reported on real validation data under different mixtures of real and PathAR-generated synthetic image--mask pairs. $\Delta$ denotes the Dice gains over the Group I baseline.
  }
    \resizebox{\textwidth}{!}{
    \begin{tabular}{ccc|cc|cc|cc|cc}
    \toprule
       \multirow{2}{*}{\textbf{Group}}&\multirow{2}{*}{\textbf{Real Samples}} & \multirow{2}{*}{\textbf{Synthetic Samples}} &  \multicolumn{8}{c}{\textbf{Downstream Segmentation} $\uparrow$}\\
       \cline{4-11}
       ~ &~ &~ & Cyto. & $\Delta$ & Fluo. & $\Delta$ & His. & $\Delta$ & Avg. & $\Delta$\\
    \midrule
      \textbf{I} &30 & 0 & 0.791 & -- & 0.917 & -- & 0.774 & -- & 0.827 & -- \\
      \textbf{II} &0 & 100 & 0.906 & $+0.115$ & 0.954 & $+0.037$ & 0.913 & $+0.139$ & 0.923 & $+0.096$\\
      \textbf{III} &30 & 100 & 0.912 & $+0.121$ & 0.957 & $+0.040$ & 0.918 & $+0.144$ & 0.929 & $+0.102$\\
      \textbf{IV} &30 & 200 & 0.915 & $+0.124$ & 0.962 & $+0.045$ & 0.943 & $+0.169$ & 0.940 & $+0.113$\\
      \textbf{V} &30 & 400 & \textbf{0.933} & $+0.142$ & \textbf{0.976} & $+0.059$ & \textbf{0.966} & $+0.192$ & \textbf{0.958} & $+0.131$\\
    \bottomrule
    \end{tabular}}
    \label{tab:result_lowdata}
\end{table*}

\begin{table}[!t]
    \centering
    \caption{Diversity evaluation on the multimodal benchmark against representative high-performing baselines.
    LPIPS and Recall are averaged over cytology, fluorescence, and histology.}
    \resizebox{0.95\linewidth}{!}{
    \begin{tabular}{l|cc}
    \toprule
    \textbf{Method} & \textbf{Avg. LPIPS} $\uparrow$ & \textbf{Avg. Recall} $\uparrow$ \\
    \midrule
    PathDiff~\cite{pathdiff} & 0.560 & 0.936 \\
    ControlNet~\cite{ControlNet} & 0.507 & 0.808 \\
    PathopixGAN~\cite{PathoGAN} & 0.526 & 0.936 \\
    DiT~\cite{DiT} & 0.556 & 0.777 \\
    LlamaGen~\cite{LLamaGen} & 0.481 & 0.748 \\
    \midrule
    \rowcolor{gray!20}
    \textbf{PathAR (ours)} & \textbf{0.603} & \textbf{0.952} \\
    \bottomrule
    \end{tabular}}
    \label{tab:diversity}
\end{table}

\paragraph{Diversity evaluation}
We further assess the diversity of generated samples on the multimodal benchmark using LPIPS and Recall.
LPIPS is computed from pairwise perceptual distances among generated samples within each modality, reflecting intra-modality visual variation.
Recall follows the improved precision-recall protocol for generative models, measuring how well generated samples cover the real data distribution in a deep feature space. For this evaluation, we compare PathAR with representative high-performing baselines from the main comparison and report the average score across the three modalities.
As shown in Table~\ref{tab:diversity}, PathAR achieves the highest LPIPS of 0.603 and Recall of 0.952, outperforming the
second-best LPIPS and Recall scores of 0.560 and 0.936, respectively. These results indicate that the improved fidelity of PathAR is accompanied by stronger intra-modality diversity and broader coverage of the real pathology distribution.

\subsection{Data Augmentation}
\label{sec:data_augmentation}
To evaluate the practical utility of PathAR-generated paired samples, we study data augmentation for downstream segmentation on the multimodal benchmark.
Following the protocol in Table~\ref{tab:result_lowdata}, we simulate a low-data regime by training a segmentation model on a limited set of 30 real labeled samples per modality.
We then progressively expand the training pool with increasing quantities of synthetic pairs (Groups I--V) and measure Dice on a fixed real validation set.
Consistent performance gains are observed as synthetic data is integrated, with the average Dice increasing from 0.827 in Group I to 0.958 in Group V.
Notably, using synthetic pairs alone already improves the average Dice from 0.827 to 0.923, indicating that the generated pairs carry segmentation-relevant structural cues even without real labeled samples.
This monotonic trend across all modalities suggests that PathAR-generated samples provide transferable structural supervision for segmentation under data-scarce conditions.

\begin{table}[!t]
    \centering
    \caption{External validation on PanNuke under organ-label conditioning. FID-DINOv2, KID-UNI, and KID-DINOv2 are averaged over all 19 organ labels.}
    \resizebox{\linewidth}{!}{
    \begin{tabular}{l|ccc}
    \toprule
    \textbf{Method} & \textbf{FID-DINOv2} $\downarrow$ & \textbf{KID-UNI} $\downarrow$ & \textbf{KID-DINOv2} $\downarrow$ \\
    \midrule
    PathDiff~\cite{pathdiff} & 175.681 & 0.535 & 0.314 \\
    ControlNet~\cite{ControlNet} & 217.918 & 0.865 & 0.644 \\
    PathopixGAN~\cite{PathoGAN} & 369.801 & 1.165 & 1.083 \\
    DiT~\cite{DiT} & 745.037 & 1.298 & 3.568 \\
    LlamaGen~\cite{LLamaGen} & 682.698 & 2.244 & 3.273 \\
    \midrule
    \rowcolor{gray!20}
    \textbf{PathAR (ours)} & \textbf{149.137} & \textbf{0.273} & \textbf{0.157} \\
    \bottomrule
    \end{tabular}}
    \label{tab:pannuke}
\end{table}

\subsection{Generalization to Organ-Label Conditioning}
\label{sec:organ_label_generation}
We further evaluate PathAR on the larger PanNuke dataset~\cite{gamper2020pannuke} under organ-label-conditioned histopathology generation, extending the conditioning variable from imaging modality to organ category.
In this setting, the condition label $c$ specifies an organ category within a single histopathology modality rather than the imaging modality.
This evaluation preserves the same label-conditioned generation formulation while testing whether PathAR can model organ-dependent morphology and appearance under finer intra-modality label variation.

\begin{figure*}[!t]
    \centering
    \includegraphics[width=\textwidth]{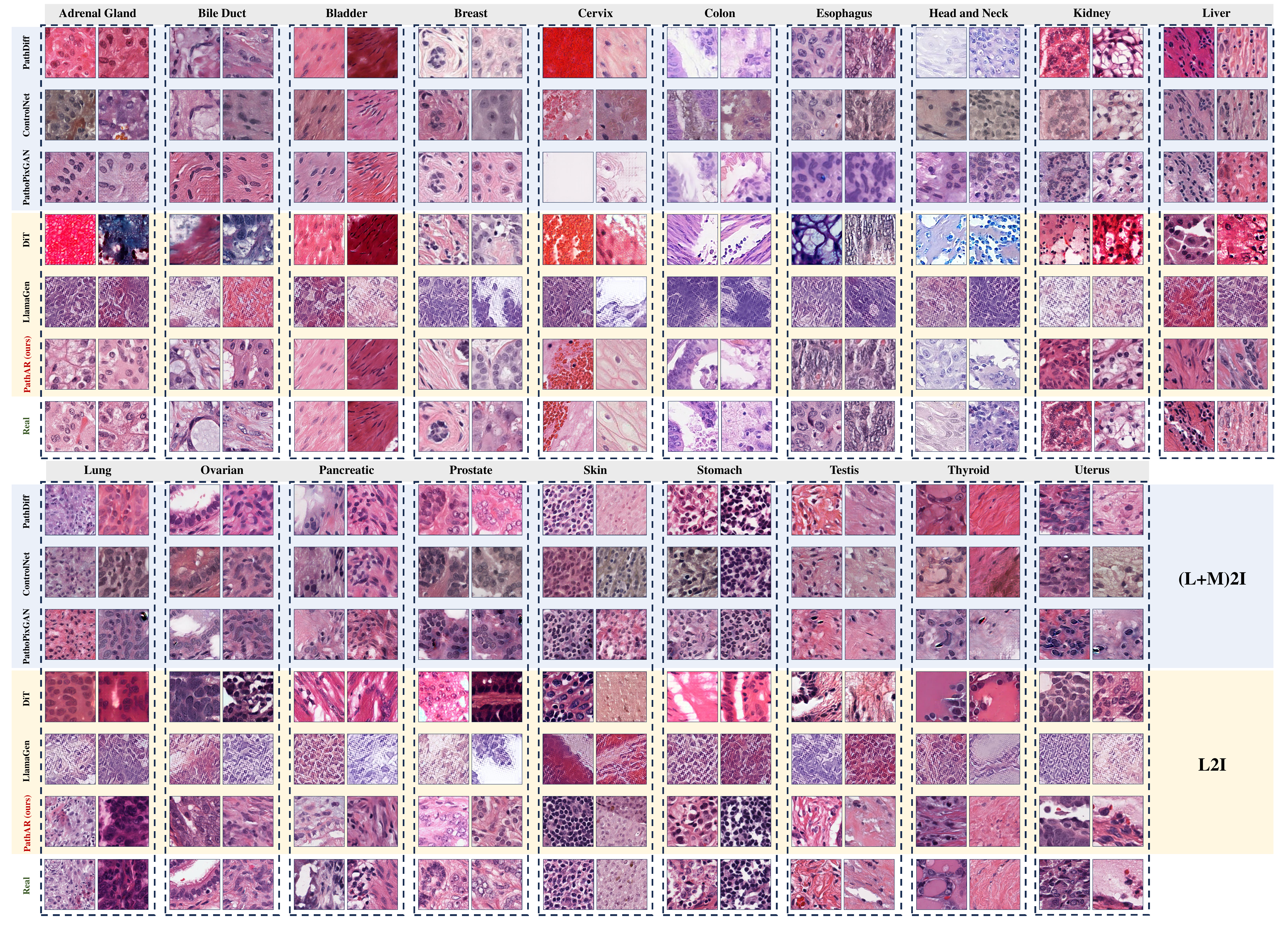}
    \caption{\textbf{Qualitative comparison on PanNuke.}
    Generated samples from representative methods are shown across the 19 organ labels of PanNuke. At inference, (L+M)2I methods are given the organ label and the ground-truth mask to generate an image, whereas L2I methods are given only the organ label to generate an image. PathAR maintains stable cellular morphology and fine-grained histological texture across diverse categories.}
    \label{fig:pannuke}
\end{figure*}

We compare PathAR with representative high-performing baselines from the primary evaluation, including pathology-specific and general generative models.
Metrics are computed over all 19 organ labels and reported as the average across labels.
As shown in Table~\ref{tab:pannuke}, PathAR achieves the best average performance among the compared methods, with FID-DINOv2 of
149.137, KID-UNI of 0.273, and KID-DINOv2 of 0.157. 
Compared with the strongest baseline PathDiff, PathAR reduces FID-DINOv2 by 26.544 and also lowers KID-UNI and KID-DINOv2 by 0.262 and 0.157, respectively.
Together with the qualitative results in Fig.~\ref{fig:pannuke}, these results indicate that PathAR remains effective when the conditioning space shifts from coarse modality labels to finer organ labels in a larger histopathology dataset.

\begin{table*}[!t]
\centering
\caption{Ablation study of PathAR components. DVQ denotes Dual-VQ tokenization. IAR denotes the complete interleaved autoregressive design, including the interleaved token sequence and asymmetric attention visibility. \textbf{Asymmetric} denotes asymmetric attention and \textbf{Multimodal} denotes joint training on all modalities.}
\resizebox{\textwidth}{!}{
\begin{tabular}{ccccc|cccc|cccc}
\toprule
\multicolumn{5}{c|}{\textbf{Component}} &
\multicolumn{4}{c|}{\textbf{FID-DINOv2} $\downarrow$} &
\multicolumn{4}{c}{\textbf{KID-UNI} $\downarrow$} \\
\cline{1-13}
\textbf{DVQ} & \textbf{IAR} & \textbf{Token Sequence} & \textbf{Asymmetric} & \textbf{Multimodal}
& Cyto. & Fluo. & His. & Avg.
& Cyto. & Fluo. & His. & Avg. \\
\midrule
$\times$ & $\times$ & -- & $\times$ & \checkmark
& 244.917 & 492.126 & 273.493 & 336.845
& 1.280 & 0.809 & 0.798 & 0.963 \\
\checkmark & $\times$ & $a_1,\ldots,a_N$ & $\times$ & \checkmark
& 222.278 & 328.742 & 212.106 & 254.376
& 0.832 & 0.399 & 0.773 & 0.668 \\
\checkmark & $\times$ & $s_1,\ldots,s_N,a_1,\ldots,a_N$ & $\times$ & \checkmark
& 395.240 & 661.750 & 446.150 & 501.050
& 1.290 & 1.380 & 2.380 & 1.680 \\
\checkmark & $\times$ & $a_1,\ldots,a_N,s_1,\ldots,s_N$ & $\times$ & \checkmark
& 267.480 & 823.430 & 233.530 & 441.480
& 1.520 & 2.355 & 1.680 & 1.850 \\
\checkmark & $\times$ & $s_1,a_1,\ldots,s_N,a_N$ & $\times$ & \checkmark
& 152.820 & 260.730 & 109.620 & 174.390
& 0.540 & 0.780 & 1.050 & 0.790 \\
\checkmark & \checkmark & $s_1,a_1,\ldots,s_N,a_N$ & \checkmark & $\times$
& 108.162 & 499.828 & 193.638 & 267.209
& 0.816 & 1.470 & 0.536 & 0.941 \\
\checkmark & \checkmark & $s_1,a_1,\ldots,s_N,a_N$ & \checkmark & \checkmark
& 128.283 & 167.541 & 95.834 & \textbf{130.552}
& 1.049 & 0.186 & 0.495 & \textbf{0.576} \\
\bottomrule
\end{tabular}}
\label{tab:ablation_study}
\end{table*}

\subsection{Ablation Study and Model Analysis}
\label{sec:ablation_analysis}
Beyond evaluating final generation quality, we further investigate the necessity of the key design choices in PathAR and examine whether the learned token streams exhibit the intended structure--appearance decomposition. Specifically, our analysis covers component effectiveness, tokenizer capacity, structure--appearance token behavior, feature-space alignment, and sampling sensitivity.

Table~\ref{tab:ablation_study} provides a component-level ablation study of PathAR, covering representation factorization, token sequence ordering, asymmetric attention, and multimodal training.
Here, variants without asymmetric attention use standard causal attention.
The first comparison evaluates the effect of Dual-VQ tokenization.
Replacing the single-codebook representation with Dual-VQ and modeling only the appearance stream reduces the average FID-DINOv2 from 336.845 to 254.376, indicating that factorized tokenization yields a more effective discrete representation for pathology images.
Nevertheless, this appearance-only prior remains substantially weaker than full PathAR, which shows that the structure stream is not merely an auxiliary training signal but an explicit generative variable required for morphology-aware synthesis.

Given the Dual-VQ representation, we next examine how structure and appearance tokens should be organized in the autoregressive prior.
The two block-ordered variants test whether the model can first generate one complete stream and then the other, while the interleaved variant preserves the local correspondence between each structure token and its spatially aligned appearance token.
The large performance gap between the block-ordered variants and the interleaved variant demonstrates that local structure--appearance coupling is essential for effective appearance rendering.
Moreover, interleaving alone is not sufficient: under the same interleaved sequence and multimodal setting, replacing standard causal attention with asymmetric attention further reduces the average FID-DINOv2 from 174.390 to 130.552 and KID-UNI from 0.790 to 0.576.
This confirms that explicitly blocking appearance-to-structure attention is critical for maintaining the intended structure-first dependency.
Finally, comparing the single-modality and multimodal variants shows that joint training improves the average performance by exploiting structural regularities shared across pathology modalities while preserving modality-specific appearance modeling.

\paragraph{Tokenizer reconstruction and capacity}
\begin{figure*}[!t]
    \centering
    \includegraphics[width=0.9\textwidth]{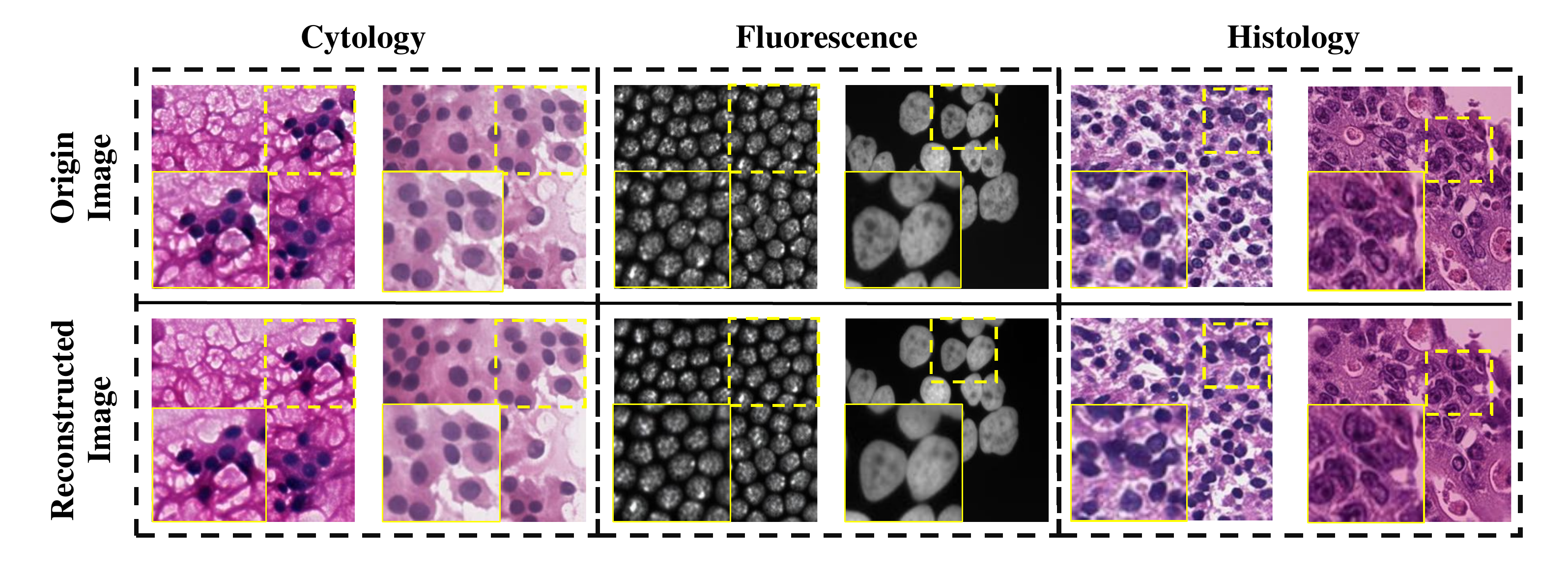}
    \caption{\textbf{Dual-VQ reconstruction examples.}
    Real pathology images and their Dual-VQ reconstructions are compared under the default tokenizer setting ($r=16$, $K_s=K_a=1024$).
    The reconstructions retain the main cellular morphology, local tissue organization, and modality-specific texture patterns, indicating that Dual-VQ preserves the key structural and appearance cues required for autoregressive generation.}
    \label{fig:vq_reconstruction}
\end{figure*}

\begin{table*}[!t]
\centering
\caption{Tokenizer hyperparameter ablation.
(A) Effect of the downsampling factor $r$, evaluated by FID-DINOv2 on cytology, fluorescence, and histology, together with their average.
(B) Effect of codebook size, evaluated by modality-averaged FID-DINOv2, KID-UNI, and codebook usage. $K_s$ and $K_a$ denote the structure and appearance codebook sizes, respectively. $S$ Usage and $A$ Usage denote the percentages of
activated entries in the structure and appearance codebooks.}
\resizebox{\textwidth}{!}{
\begin{tabular}{c|cccc||c|cccc}
\toprule
\multicolumn{5}{c||}{\textbf{(A) Downsampling Factor (FID-DINOv2)}  } &
\multicolumn{5}{c}{\textbf{(B) Codebook Size}} \\
\midrule
\textbf{$r$} & \textbf{Cyto.} & \textbf{Fluo.} & \textbf{His.} & \textbf{Avg.} &
\textbf{$K_s=K_a$} & \textbf{Avg. FID-DINOv2} $\downarrow$ & \textbf{Avg. KID-UNI} $\downarrow$ & \textbf{$S$ Usage} $\uparrow$ & \textbf{$A$ Usage} $\uparrow$ \\
\midrule
32 & 183.080 & 198.960 & 361.790 & 247.940 &
4096 & 587.079 & 1.858 & 17.11\% & 17.16\% \\
16 & 128.283 & \textbf{167.541} & \textbf{95.834} & \textbf{130.552} &
2048 & 252.950 & 0.883 & 31.05\% & 31.39\% \\
8 & \textbf{103.440} & 314.030 & 129.300 & 182.260 &
1024 & \textbf{130.552} & \textbf{0.576} & \textbf{93.07\%} & \textbf{93.16\%} \\
\bottomrule
\end{tabular}}
\label{tab:tokenizer_ablation}
\end{table*}

The reconstruction examples in Fig.~\ref{fig:vq_reconstruction} show the visual fidelity of Dual-VQ under the default setting.
Across modalities, the reconstructed images preserve the main cell layouts and tissue organization observed in the original samples, while also retaining modality-specific texture patterns such as staining appearance and fluorescence intensity.
These results indicate that Dual-VQ provides a high-fidelity discrete representation for subsequent autoregressive modeling by preserving the structural and appearance cues required for generation, rather than discarding fine-grained pathology-relevant information during tokenization.

\begin{figure*}[!t]
    \centering
    \includegraphics[width=0.93\textwidth]{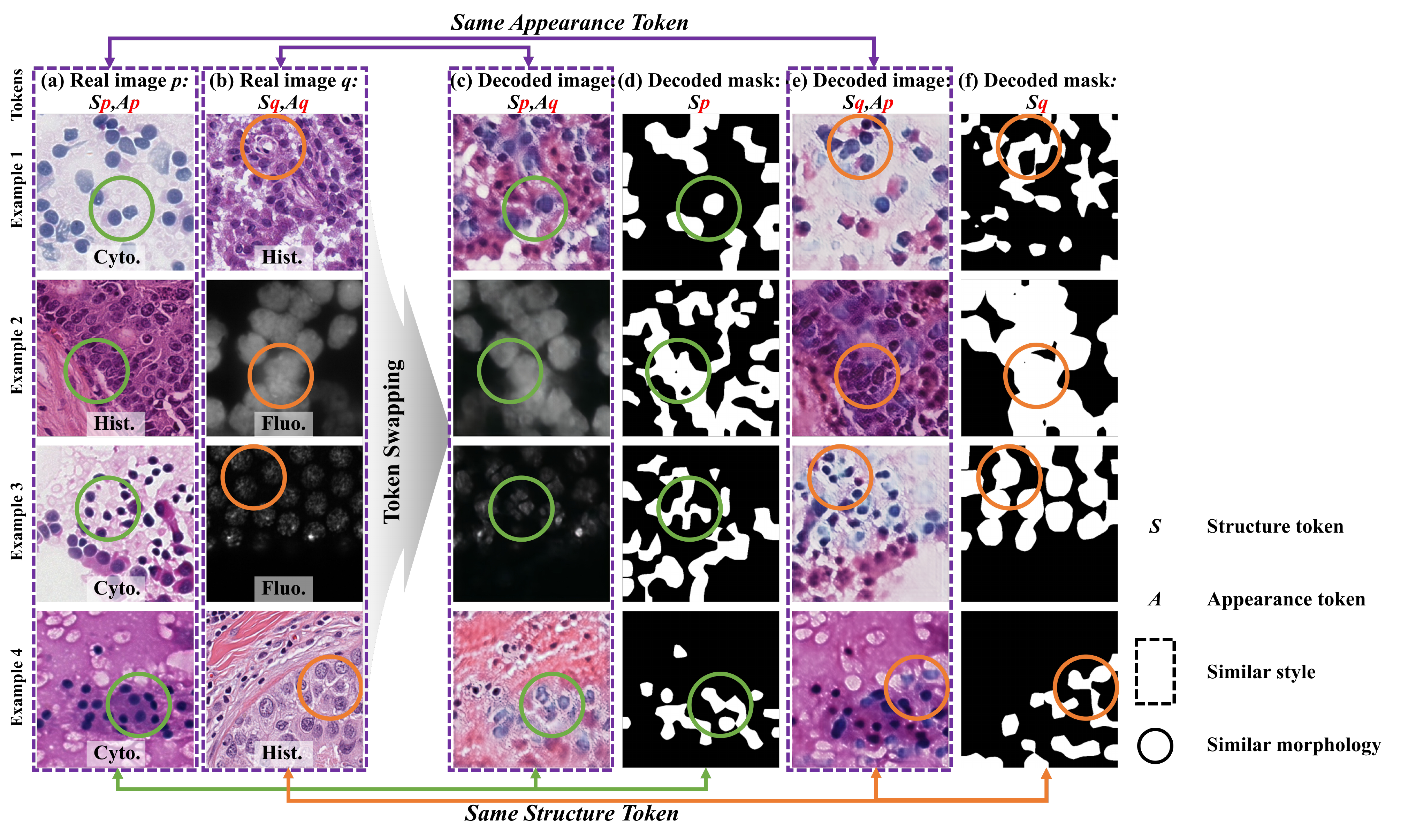}
    \caption{\textbf{Cross-sample swapping of structure and appearance tokens across modalities.}
    Each row presents one token-swapping example constructed from two real samples of different modalities.
    (a) and (b) denote two real images indexed by $p$ and $q$, which are encoded into structure--appearance token pairs $(S_p,A_p)$ and $(S_q,A_q)$, respectively, where $S$ and $A$ denote the structure and appearance token streams. After token swapping, (c) and (e) show decoded images from $(S_p,A_q)$ and $(S_q,A_p)$, respectively, while (d) and (f) show decoded masks from $S_p$ and $S_q$, respectively.
    Purple arrows indicate shared appearance tokens, and orange or green arrows indicate shared structure tokens.
    The highlighted regions show that morphology is primarily determined by the structure stream, while style and modality are primarily determined by the appearance stream.}
    \label{fig:sa_swap}
\end{figure*}

We further evaluate tokenizer hyperparameters that control the resolution and capacity of the discrete latent representation.
As shown in Table~\ref{tab:tokenizer_ablation} (A), the downsampling factor $r=16$ achieves the best average FID-DINOv2 across the three modalities.
Reducing the downsampling factor to $r=8$ increases the latent grid resolution, but also substantially lengthens the autoregressive sequence; although it improves cytology, it degrades fluorescence and histology, suggesting that higher token resolution does not necessarily translate into better generation quality under the current data scale.
Conversely, increasing the downsampling factor to $r=32$ overly compresses the latent grid, which weakens the preservation of fine-grained morphology and leads to inferior performance, particularly on histology.
These results indicate that $r=16$ provides a practical balance between structural detail retention and autoregressive modeling complexity.

Table~\ref{tab:tokenizer_ablation} (B) further evaluates the codebook capacity of the Dual-VQ tokenizer.
The best performance is achieved with $K_s=K_a=1024$, while increasing the codebook size to 2048 and 4096 degrades the average FID-DINOv2 from 130.552 to 252.950 and 587.079, respectively.
The average KID-UNI follows the same trend, increasing from 0.576 to 0.883 and 1.858.
Meanwhile, the usage rates of both structure and appearance codebooks drop from over 93\% at 1024 entries to about 31\% and 17\% at 2048 and 4096 entries.
These results indicate that, under the current pathology data scale, larger codebooks introduce under-utilized entries rather than improving the learned discrete representation.

\paragraph{Structure--appearance token analysis}
To examine whether the structure and appearance streams are functionally separated in modeling morphology and modality-specific appearance, we perform cross-sample token swapping between real samples from different modalities.
Given two different images $p$ and $q$, the Dual-VQ tokenizer encodes them into structure--appearance pairs $(S_p,A_p)$ and $(S_q,A_q)$. We then decode images from the swapped pairs $(S_p,A_q)$ and $(S_q,A_p)$, while decoding masks from swapped $S_p$ and $S_q$ to visualize the structural information carried by each structure stream.
If the learned factorization is effective, the decoded image from $(S_p,A_q)$ should preserve the morphological layout associated with $S_p$ while adopting the modality-specific appearance associated with $A_q$, and vice versa for $(S_q,A_p)$.
As shown in Fig.~\ref{fig:sa_swap}, images sharing the same structure tokens preserve similar cellular organization and mask geometry, whereas images sharing the same appearance tokens exhibit similar staining, color, and texture statistics, and modality-specific visual patterns.
As cross-modality token recombination may yield unseen structure--appearance pairings, the analysis focuses on factor preservation and transfer.
The highlighted regions further show that local morphology is primarily governed by the structure stream, while visual style is primarily governed by the appearance stream.
These results support the intended functional separation between morphological layout and modality-specific appearance, without assuming complete independence between the two factors.

\begin{table}[!t]
    \centering
    \caption{Pairwise Euclidean distances between modality centroids in the token embedding space.
    Centroids are computed from real samples encoded by the trained Dual-VQ tokenizer. Structure and appearance distances are computed between centroids of structure-token and appearance-token embeddings, respectively. Lower cross-modal distances indicate stronger modality invariance.}
    \resizebox{\linewidth}{!}{
    \begin{tabular}{l|cc}
    \toprule
    \textbf{Modality Pair} & \textbf{Structure Distance} & \textbf{Appearance Distance} \\
    \midrule
    Hist.--Fluo. & 1.861 & 3.743 \\
    Hist.--Cyto. & 1.303 & 1.495 \\
    Fluo.--Cyto. & 3.090 & 5.237 \\
    \midrule
    Avg. $\pm$ Std. & 2.085$\pm$0.746 & 3.492$\pm$1.538 \\
    \bottomrule
    \end{tabular}}
    \label{tab:token_distance}
\end{table}

To further characterize this separation, we measure cross-modal distances in the token embedding space.
For each modality and each stream, real samples are encoded by the trained Dual-VQ tokenizer, and a modality centroid is computed by averaging the resulting token embeddings over all samples and spatial locations.
Pairwise Euclidean distances are then calculated between modality centroids within the structure and appearance streams.
As shown in Table~\ref{tab:token_distance}, structure-token centroids are consistently closer across modalities than appearance-token centroids.
Across the three modality pairs, the mean distances of the structure and appearance streams are $2.085$ and $3.492$, with standard deviations of $0.746$ and $1.538$, respectively.
These smaller structure-stream distances indicate that structural representations from different modalities occupy a more compact shared embedding region, whereas the larger appearance-stream distances reflect stronger modality-dependent separation.
This pattern is consistent with the intended factorization: the structure stream captures morphology shared across modalities, while the appearance stream preserves modality-specific visual statistics.

\paragraph{Feature-space visualization and sampling hyperparameters}
\begin{figure*}[!t]
    \centering
    \includegraphics[width=\textwidth]{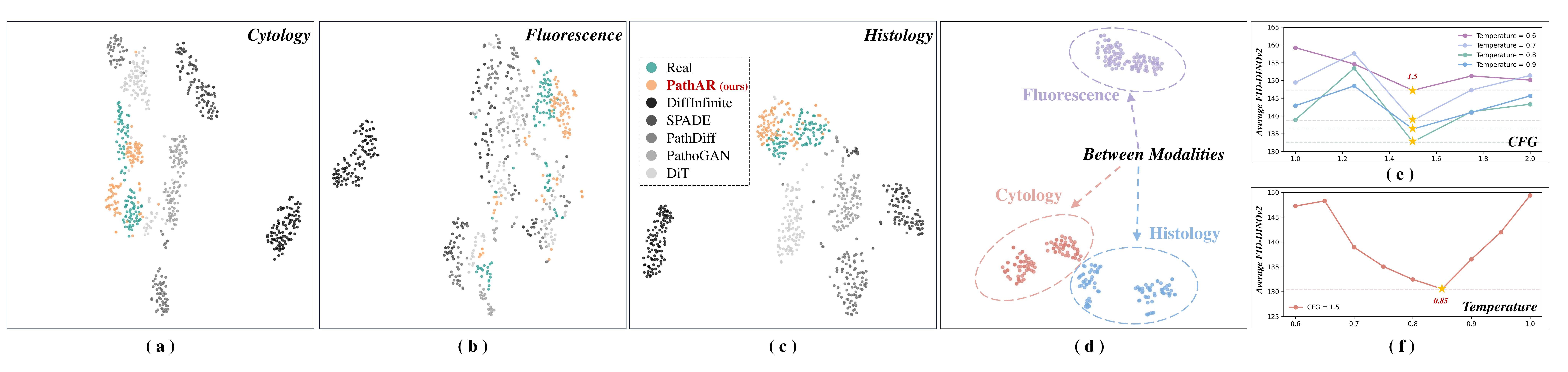}
    \caption{\textbf{t-SNE analysis of generated images and sampling hyperparameters.}
    (a--d) \textbf{Feature t-SNE.} t-SNE visualizations of UNI embeddings for real and generated images.
    (a--c) Per-modality alignment: PathAR (\textcolor[RGB]{245,169,109}{orange}) closely overlaps with Real (\textcolor[RGB]{74,171,159}{green}) distributions.
    (d) Joint embedding of PathAR samples demonstrates distinct inter-modality separation and rich intra-modality diversity.
    (e--f) \textbf{Hyperparameter tuning.} Impact of CFG scale and temperature on average FID-DINOv2. Optimal performance is achieved at CFG $=1.5$ and temperature $=0.85$ (with fixed top-$k=100$, top-$p=0.92$).}
    \label{fig:tsne}
\end{figure*}

Fig.~\ref{fig:tsne} presents t-SNE visualizations of UNI~\cite{uni} feature embeddings of real and generated samples.
In the per-modality plots, panels (a)--(c), PathAR samples lie closer to the real clusters than baselines, indicating reduced embedding-space shift.
The embedding gaps are smaller in fluorescence due to its relatively homogeneous appearance, while the gap becomes more apparent for cytology and histology, where baselines deviate more from the real clusters.
In the joint embedding, panel (d), PathAR preserves clear modality separation while retaining intra-modality diversity.
Cytology and histology form adjacent yet separable clusters, which is consistent with shared visual patterns under the UNI representation.
Sampling sensitivity is evaluated in Fig.~\ref{fig:tsne}(e) and (f). The CFG scale of $1.5$ and temperature of $0.85$ achieve the best average FID-DINOv2 across the three modalities.

\subsection{Computational Efficiency}
\label{sec:computational_efficiency}

\begin{table}[!t] 
    \centering
    \caption{Inference time and parameter count for AR-based models at $256\times256$ resolution. The best results are shown in \textbf{bold}. The second-best results are \underline{underlined}.}
    \resizebox{\linewidth}{!}{
    \begin{tabular}{l|cc}
    \toprule
    \textbf{Method} & \textbf{Inference Time (s / image)} $\downarrow$ & \textbf{Params (M)} $\downarrow$ \\
    \midrule
       LlamaGen~\cite{LLamaGen}     & 10.35 & \underline{182.77}
\\
       VAR~\cite{VAR}               & 0.28 & 419.23
\\
       ControlAR~\cite{controlar}   & 3.53 & 222.01
\\
       ImageFolder~\cite{imagefolder} & \textbf{0.077} & 572.30
\\
    \midrule
    \rowcolor{gray!20}
      \textbf{PathAR (ours)}        & \underline{0.086} & \textbf{170.73} \\
    \bottomrule
    \end{tabular}}
    \label{tab:efficiency_ar}
\end{table}

Table~\ref{tab:efficiency_ar} quantifies the computational overhead at $256 \times 256$ resolution in terms of inference latency and parameter count.
PathAR demonstrates the smallest parameter count ($170.73$M) among evaluated autoregressive benchmarks, while maintaining a competitive inference latency of $0.086$ s per image.
These results show that PathAR achieves competitive inference speed with the smallest parameter count among the AR-based baselines, indicating a favorable trade-off between generation quality and computational cost.

\section{Conclusion}
\label{sec:conclusion}
In this work, we present PathAR, a structure-first multimodal autoregressive framework that explicitly factorizes biological structure and appearance.
By enforcing a structure-to-appearance generation process, PathAR produces interpretable segmentation masks alongside high-fidelity images, enabling controllable and structurally grounded pathology synthesis.
Experiments on the primary multimodal benchmark show that PathAR improves image fidelity, modality-specific appearance consistency, structural accuracy, sample diversity, and downstream segmentation utility. The PanNuke evaluation further demonstrates that the same factorized conditional generation framework remains effective under finer intra-modality organ-label variation.
Despite these strengths, the current study focuses on patch-level generation, while whole-slide pathology synthesis remains challenging due to gigapixel-scale spatial context and memory demands.
Future work will explore scalable generation strategies for higher-resolution and whole-slide pathology synthesis, further extending structure-aware generative modeling to broader pathology distributions.

\bibliographystyle{IEEEtran}
\bibliography{Main}

\newpage

\section{Biography Section}
  \begin{IEEEbiographynophoto}{Yuan Zhang}
  Yuan Zhang received the B.S. degree in 2019 from Northwestern Polytechnical University, Xi'an, China, and the M.S. degree in 2020 from Imperial College London, London, U.K. She is currently pursuing the Ph.D. degree in computer science and technology with Southeast University, Nanjing, China. Her research interests include
  medical image analysis, pathology image synthesis, and multimodal learning.
  \end{IEEEbiographynophoto}

  \begin{IEEEbiographynophoto}{Jiahao Xia}
  Jiahao Xia received the B.S. degree in software engineering from Central South University, Hunan, China. He received the M.S. degree in computer science and technology from Southeast University, Nanjing, China. His research interests include
  computer vision, medical image analysis, and artificial intelligence-generated content.
  \end{IEEEbiographynophoto}

  \begin{IEEEbiographynophoto}{Junzhang Huang}
  Junzhang Huang is currently pursuing the Ph.D. degree in computer science and technology with Southeast University, Nanjing, China. His research interests include medical image analysis.
  \end{IEEEbiographynophoto}

  \begin{IEEEbiographynophoto}{Meng Wang}
  Meng Wang received the Ph.D. degree from Soochow University, Suzhou, China, in 2021. He is an Assistant Professor at the Centre for Innovation \& Precision Eye Health and Department of Ophthalmology, Yong Loo Lin School of Medicine, National University of Singapore. His research interests include precision and trustworthy artificial intelligence for healthcare.
  \end{IEEEbiographynophoto}

  \begin{IEEEbiographynophoto}{Feng Chen}
  Feng Chen is a Professor with the Department of Biostatistics, Center for Global Health, School of Public Health, Nanjing Medical University, Nanjing, China. He previously served as the Dean of the School of Public Health and the Dean of the Graduate School at Nanjing Medical University. His research interests include biostatistics, medical data analysis, and artificial intelligence for healthcare.
  \end{IEEEbiographynophoto}

  \begin{IEEEbiographynophoto}{Guanyu Yang}
  Guanyu Yang (Senior Member, IEEE) received the B.S. and M.S. degrees in biomedical engineering from Southeast University, Nanjing, China, in 2002 and 2005, respectively, and the Ph.D. degree in signal and image processing from Leiden University Medical Center, Leiden University, Leiden, The Netherlands, in 2008. From 2009 to 2011, he held a postdoctoral fellowship at Leiden University. In 2011, he joined Southeast University, where he is currently the Vice Dean of the School of Computer Science and Engineering, a Professor, a Doctoral Supervisor, and the Head of the Department of Imaging Science and Technology. In 2018, he served as a member of the Medical Imaging Professional Committee of the Chinese Society of Graphics. His research interests include medical image analysis, biomedical image processing, and artificial intelligence for healthcare.
  \end{IEEEbiographynophoto}

  \begin{IEEEbiographynophoto}{Huazhu Fu}
  Huazhu Fu (Senior Member, IEEE) received the Ph.D. degree from Tianjin University, Tianjin, China, in 2013. He is currently a Principal Scientist with the Institute of High Performance Computing, Agency for Science, Technology and Research (A*STAR),
  Singapore. His research interests include medical image analysis, artificial intelligence for healthcare, and trustworthy AI. He has been recognized as a Highly Cited Researcher by Clarivate and among the Top 2\% Scientists Worldwide by Stanford University. He serves as an Associate Editor for several IEEE journals, including IEEE Transactions on Medical Imaging, IEEE Transactions on Neural Networks and Learning Systems, and IEEE Journal of Biomedical and Health Informatics.
  \end{IEEEbiographynophoto}

 




\vfill

\end{document}